\documentclass{article} 
\usepackage[accepted]{icml2014}
\usepackage{times}
\usepackage{natbib}

\usepackage{notations}
\usepackage{myAlgorithm}
\renewcommand{\algotab}[0]{\hspace*{2\labelsep}}

\usepackage[english]{babel}
\usepackage{bm,amsmath,amssymb,accents}
\usepackage{graphicx}
\usepackage{subfigure}
\usepackage{units}
\usepackage{xspace}
\usepackage{multirow}
\usepackage{tablefootnote}
\usepackage{booktabs}

\usepackage{graphicx}
\usepackage{color}
\definecolor{darkblue}{rgb}{0,0,0.5}
\definecolor{firebrick}{rgb}{0.75,0.125,0.125}
\definecolor{darkgreen}{rgb}{0,0.5,0}
\usepackage[colorlinks=true,linkcolor=firebrick,citecolor=darkgreen,urlcolor=darkblue]{hyperref}
\usepackage%
{trackchanges}
\renewcommand{\initialsTwo}{BK}
\renewcommand{\initialsThree}{RBF}

\newcommand{\sectionBefore}{0pt}
\newcommand{\sectionAfter}{0pt}

\definecolor{darkred}{rgb}{.70,0,0}

\icmltitlerunning{Multi-class Hamming trees}

\begin{document}

\twocolumn[ 

\icmltitle{The return of \Algo{AdaBoost.MH}: multi-class Hamming trees}

\icmlauthor{Bal\'azs K\'egl}{balazs.kegl@gmail.com}
\icmladdress{LAL/LRI, University of Paris-Sud, CNRS, 91898
Orsay, France}

\icmlkeywords{boring formatting information, machine learning, ICML}

\vskip 0.3in
]

\newcommand{\fix}{\marginpar{FIX}}
\newcommand{\new}{\marginpar{NEW}}


\begin{abstract}
Within the framework of \Algo{AdaBoost.MH}, we propose to train vector-valued
decision trees to optimize the multi-class edge without reducing the multi-class
problem to $K$ binary one-against-all classifications. The key element of the
method is a vector-valued decision stump, factorized into an input-independent
vector of length $K$ and label-independent scalar classifier. At inner tree
nodes, the label-dependent vector is discarded and the binary classifier can be
used for partitioning the input space into two regions. The algorithm retains
the conceptual elegance, power, and computational efficiency of binary
\Algo{AdaBoost}. In experiments it is on par with support vector machines and
with the best existing multi-class boosting algorithm \Algo{AOSOLogitBoost}, and
it is significantly better than other known implementations of
\Algo{AdaBoost.MH}.
\end{abstract}

\vspace{\sectionBefore}
\section{Introduction}
\vspace{\sectionAfter}

\Algo{AdaBoost}~\cite{FrSc97} is one of the most influential supervised learning
algorithms of the last twenty years. It has inspired learning theoretical
developments and also provided a simple and easily interpretable modeling tool
that proved to be successful in many applications~\cite{CaNi06}. It is
especially the method of choice when any-time solutions are required on large
data sets, so it has been one of the most successful techniques in recent
large-scale classification and ranking challenges \cite{KDDCup09,ChChLi11a}.

The original \Algo{AdaBoost} paper of Freund and Schapire~\cite{FrSc97}, besides
defining binary \Algo{AdaBoost}, also described two multi-class extensions,
\Algo{AdaBoost.M1} and \Algo{AdaBoost.M2}. Both required a quite strong
performance from the base learners, partly defeating the purpose of boosting,
and saw limited practical success. The breakthrough came with Schapire and
Singer's seminal paper~\cite{ScSi99}, which proposed, among other interesting
extensions, \Algo{AdaBoost.MH}. The main idea of the this approach is to use
\emph{vector}-valued base classifiers to build a multi-class discriminant
function of $K$ outputs (for $K$-class classification). The weight vector, which
plays a crucial role in binary \Algo{AdaBoost}, is replaced by a weight
\emph{matrix} over instances and labels. The simplest implementation of the
concept is to use $K$ independent one-against-all classifiers in which base
classifiers are only loosely connected through the common normalization of the
weight matrix. This setup works well with single decision stumps, but in most of
the practical problems, boosting stumps is suboptimal compared to boosting more
complex base classifiers such as trees. Technically, it is possible to build $K$
one-against-all binary decision trees in each iteration, but this approach, for
one reason or another, has not produced state-of-the-art results. As a
consequence, several recent papers concentrate on replacing the boosting
objective and the engine that optimizes this
objective~\cite{Li09,Li09a,ZhZoRoHa09,SuReZh12,MuSc13}.

The main misconception that comes back in several papers is that
\Algo{AdaBoost.MH} \emph{has to} train $K$ parallel one-against-all classifiers
in each iteration. It turns out that the original setup is more general. For
example, staying within the classical \Algo{AdaBoost.MH} framework,
\citet{KeBu09} trained products of simple classifiers and obtained
state-of-the-art results on several data sets. In this paper, we describe
\emph{multi-class Hamming trees}, another base learner that optimizes the
multi-class edge without reducing the problem to $K$ binary classifications. The
key idea is to \emph{factorize} general vector-valued classifiers into an
input-independent vector of length $K$, and label-independent \emph{scalar}
classifier. It turns out that optimizing such base classifiers using decision
stumps as the scalar component is almost as simple as optimizing simple binary
stumps on binary data. The technique can be intuitively understood as optimizing
a binary cut and an output code at the same time. The main consequence of the
setup is that now it is easy to build \emph{trees} of these classifiers by
simply discarding the label-dependent vector and using the binary classifier for
partitioning the input space into two regions.

The algorithm retains the conceptual elegance, power, and computational
efficiency of binary \Algo{AdaBoost}. Algorithmically it cannot fail (the edge
is always positive) and in practice it almost never overfits. Inheriting the
flexibility of \Algo{AdaBoost.MH}, it can be applied directly (without any
modification) to multi-label and multi-task classification. In experiments
(carried out using an open source package of \citet{BBCCK11} for
reproducibility) we found that \Algo{AdaBoost.MH} with Hamming trees performs on
par with the best existing multiclass boosting algorithm
\Algo{AOSOLogitBoost}~\cite{SuReZh12} and with support vector machines (SVMs;
\citealt{BoGuVa92}). It is also significantly better than other known
implementations of \Algo{AdaBoost.MH}~\cite{ZhZoRoHa09,MuSc13}.

The paper is organized as follows. In Section~\ref{secAdaBoostMH} we give the
formal multi-class setup used in the paper and \Algo{AdaBoost.MH}, and show how
to train factorized base learners in general. The algorithm to build Hamming
trees is described in Section~\ref{secHammingTrees}. Experiments are described
in Section~\ref{secExperiments} before a brief conclusion in
Section~\ref{secConclusion}.

\vspace{\sectionBefore}
\section{\Algo{AdaBoost.MH}}
\vspace{\sectionAfter}
\label{secAdaBoostMH}

In this section we first introduce the general multi-class learning setup
(Section~\ref{secBoostingPreliminaries}), then we describe \Algo{AdaBoost.MH} in
detail (Section~\ref{secAdaBoostMHSub}). We proceed by explaining the general
requirements for base learning in \Algo{AdaBoost.MH}, and introduce the notion
of the factorized vector-valued base learner
(Section~\ref{secBaseLearning}). Finally, we explain the general objective for
factorized base learners and the algorithmic setup to optimize that objective.
(Section~\ref{secCastingTheVotes}).

\subsection{The multi-class setup: single-label and multi-label/multi-task}
\label{secBoostingPreliminaries}

For the formal description of \Algo{AdaBoost.MH}, let the training data be $\cD
= \big\{(\bx_1,\by_1), \ldots, (\bx_n,\by_n)\big\}$, where $\bx_i \in \RR^d$ are
\emph{observation} vectors, and $\by_i \in \{\pm 1\}^K$ are \emph{label}
vectors. Sometimes we will use the notion of an $n \times d$ \emph{observation
  matrix} of $\bX = (\bx_1,\ldots,\bx_n)$ and an $n \times K$ \emph{label
  matrix} $\bY = (\by_1,\ldots,\by_n)$ instead of the set of pairs
$\cD$.\footnote{We will use bold capitals $\bX$ for matrices, bold small letters
  $\bx_i$ and $\bx_{,.j}$ for its row and column vectors, respectively, and
  italic for its elements $x_{i,j}$.} In \emph{multi-class} classification, the
single label $\ell(\bx)$ of the observation $\bx$ comes from a finite
set. Without loss of generality, we will suppose that $\ell \in \cL =
\{1,\ldots,K\}$. The label vector $\by$ is a \emph{one-hot} representation of
the correct class: the $\ell(\bx)$\/th element of $\by$ will be $1$ and all the
other elements will be $-1$. Besides expressing faithfully the architecture of a
multi-class neural network or multi-class \Algo{AdaBoost}, this representation
has the advantage to be generalizable to \emph{multi-label} or \emph{multi-task}
learning when an observation $\bx$ can belong to several classes. To avoid
confusion, from now on we will call $\by$ and $\ell$ the label and the label
\emph{index} of $\bx$, respectively. For emphasizing the distinction between
multi-class and multi-label classification, we will use the term
\emph{single-label} for the classical multi-class setup, and reserve multi-class
to situations when we talk about the three setups in general.

The goal of learning is to infer a \emph{vector}-valued multi-class
\emph{discriminant} function $\f: \cX \rightarrow \RR^K$.\footnote{Instead of
  the original notation of \cite{ScSi99} where both $\bx$ and $\ell$ are inputs
  of a function $f(\bx,\ell)$ outputting a single \emph{real}-valued score, we
  use the notation $\f(\bx) = (f_1(\bx),\ldots,f_K(\bx)\big)$ since we feel it
  expresses better that $\bx$ is (in general) continuous and $\ell$ is a
  discrete index.}  The single-label output of the algorithm is then
$\ell_\f(\bx) = \argmax_\ell f_\ell(\bx)$. The classical measure of the
performance of the multi-class discriminant function $\f$ is the
\emph{single-label one-loss} $L_\I\big(\f,(\bx,\ell)) = \IND{\ell \not=
  \ell_{\f}(\bx_i)}$, which defines the single-label training error
\begin{equation}\label{eqnOneError}
\widehat{R}_\I(\f) = \frac{1}{n}\sum_{i=1}^n \IND{\ell(\bx_i) \not=
  \ell_{\f}(\bx_i)}.\footnote{The indicator function $\IND{A}$ is $1$ if its
  argument $A$ is true and $0$ otherwise.}
\end{equation}
Another, perhaps more comprehensive, way to measure the performance of $\f$
is by computing the \emph{weighted Hamming loss}
$L_{\Algo{h}}\big(\f,(\bx,\by),\bw\big) = \sum_{\ell=1}^K w_{\ell}
\IND{\sign\big(f_\ell(\bx)\big) \not= y_{\ell}}$ where $\bw =
\big[w_{\ell}\big]$ is an $\RR^K$-valued ``user-defined'' weight vector over
labels. The corresponding empirical risk (training error) is
\begin{equation}\label{eqnHammingLoss}
\widehat{R}_{\Algo{h}}(\f,\bW) = \frac{1}{n}\sum_{i=1}^n \sum_{\ell=1}^K w_{i,\ell}
\IND{\sign\big(f_\ell(\bx_i)\big) \not= y_{i,\ell}},
\end{equation}
where $\bW = \big[w_{i,\ell}\big]$ is an $n \times k$ weight matrix over data
points and labels.

In the multi-label/multi-task setup, when, for example, it is equally important
to predict that a song is ``folk'' as predicting that it is sung by a woman, the
Hamming loss with uniform weights $w_{\ell} = 1 / K, \; \ell=1,\ldots,K$ is a
natural measure of performance: it represents the uniform error rate of missing
any class sign $y_\ell$ of a given observation $\bx$. In single-label
classification, $\bw$ is usually set asymmetrically to
\begin{equation}\label{eqnMulticlassInitialWeights}
w_{\ell} = \begin{cases} \frac{1}{2} & \mbox{ if $\ell = \ell(\bx)$ (i.e., if
    $y_{\ell} = 1$),} \\ \frac{1}{2(K-1)} & \mbox{ otherwise (i.e., if
    $y_{\ell} = -1$)}.
\end{cases}
\end{equation}
The idea behind this scheme is that it will create $K$ well-balanced
\emph{one-against-all} binary classification problems: if we start with a
balanced single-label multi-class problem, that is, if each of the $K$ classes
have $n/K$ examples in $\cD$, then for each class $\ell$, the sum of the weights
of the positive examples in the column $\bw_{\cdot,\ell}$ of the weight matrix
$\bW$ will be equal to the sum of the weights of the negative examples. Note
that both schemes boil down to the classical uniform weighting in binary
classification.

\subsection{$\Algo{AdaBoost.MH}$}
\label{secAdaBoostMHSub}

The goal of the $\Algo{AdaBoost.MH}$ algorithm (\citealt{ScSi99};
Figure~\ref{figAdaBoostMH}) is to return a vector-valued discriminant function
$\f^{(T)}:\RR^d \rightarrow \RR^K$ with a small Hamming loss
$\widehat{R}_{\Algo{h}}(\f,\bW)$ (\ref{eqnHammingLoss}) by minimizing the
\emph{weighted multi-class exponential margin-based error}
\begin{equation}\label{eqnExpLoss}
\widehat{R}_{\Algo{exp}}\big(\f^{(T)},\bW\big) = \frac{1}{n}\sum_{i=1}^n
\sum_{\ell=1}^K w_{i,\ell} \exp\big(\!\!-\!f^{(T)}_\ell(\bx_i)y_{i,\ell}\big).
\end{equation}
Since $\exp(-\rho) \ge \IND{\rho < 0}$, (\ref{eqnExpLoss}) upper bounds the
Hamming loss $\widehat{R}_{\Algo{h}}\big(\f^{(T)},\bW\big)$
(\ref{eqnHammingLoss}). $\Algo{AdaBoost.MH}$ builds the final discriminant
function $\f^{(T)}(\bx) = \sum_{t=1}^T \bh^{(t)}(\bx)$ as a sum of $T$ {\em base
  classifiers} $\bh^{(t)}: \cX \rightarrow \RR^K$ returned by a {\em base
  learner} algorithm $\Algo{Base}\big(\bX,\bY,\bW^{(t)}\big)$ in each iteration
$t$.

\setlength{\algowidth}{1.0\columnwidth}

\begin{figure}[!ht]
\centerline{
\begin{algorithm}{$\Algo{AdaBoost.MH}(\bX,\bY,\bW,\Algo{Base}(\cdot,\cdot,\cdot),T)$}
\Aitem $\bW^{(1)} \setto \frac{1}{n} \bW$\label{linWeightInitialization}
\Aitem \For $t \setto 1$ \To $T$
\Aitem \mt $\big(\alpha^{(t)},\bv^{(t)},\varphi^{(t)}(\cdot)\big) \setto
\Algo{Base}\big(\bX,\bY,\bW^{(t)}\big)$
\Aitem \mt $\bh^{(t)}(\cdot) \setto \alpha^{(t)}\bv^{(t)}\varphi^{(t)}(\cdot)$
\Aitem \mt \For $i \setto 1$ \To $n$ \For $\ell \setto 1$ \To $K$
\Aitem \mtt $\displaystyle w_{i,\ell}^{(t+1)} \setto w_{i,\ell}^{(t)}
             \underbrace{\frac{e^{\!-\!h_\ell^{(t)}(\bx_i)
                 y_{i,\ell}}}{\displaystyle \sum_{i^\prime=1}^n
           \sum_{\ell^\prime=1}^K w_{i^\prime,\ell^\prime}^{(t)}
           e^{\!-\!h_{\ell^\prime}^{(t)}(\bx_{i^\prime})
               y_{i^\prime,\ell^\prime}}}}_{\displaystyle Z\big(\bh^{(t)},\bW^{(t)}\big)}$ \label{linWeightUpdate}
\Aitem \Return $\f^{(T)}(\cdot) = \sum_{t=1}^T \bh^{(t)}(\cdot)$
\end{algorithm}
}
\caption{The pseudocode of the $\Algo{AdaBoost.MH}$ algorithm with factorized
  base classifiers (\ref{eqnMulticlassBaseClassifier}). $\bX$ is the $n \times
  d$ observation matrix, $\bY$ is the $n \times K$ label matrix, $\bW$ is the
  user-defined weight matrix used in the definition of the weighted Hamming
  error (\ref{eqnHammingLoss}) and the weighted exponential margin-based error
  (\ref{eqnExpLoss}), $\Algo{Base}(\cdot,\cdot,\cdot)$ is the base learner
  algorithm, and $T$ is the number of iterations. $\alpha^{(t)}$ is the base
  coefficient, $\bv^{(t)}$ is the vote vector, $\varphi^{(t)}(\cdot)$ is the
  scalar base (weak) classifier, $\bh^{(t)}(\cdot)$ is the vector-valued base
  classifier, and $\f^{(T)}(\cdot)$ is the final (strong) discriminant function.
\label{figAdaBoostMH}}
\end{figure}

\vspace{\sectionBefore}
\subsection{Base learning for \Algo{AdaBoost.MH}}
\vspace{\sectionAfter}
\label{secBaseLearning}

The goal of multi-class base learning is to minimize the \emph{base objective}
\begin{equation}\label{eqnEnergy}
Z^{(t)} = \min_{\bh} Z\big(\bh,\bW^{(t)}\big) = \sum_{i=1}^n \sum_{\ell=1}^K
w_{i,\ell}^{(t)} e^{-h_\ell(\bx_i) y_{i,\ell}}.
\end{equation}
It is easy to show~\cite{ScSi99} that i) the one-error
$\widehat{R}_\I(\f^{(T)})$ (\ref{eqnOneError}) is upper bounded by
$\prod_{t=1}^T Z^{(t)}$, and so ii) if the standard weak-learning condition
$Z^{(t)} \le 1 - \delta$ holds, $\widehat{R}_\I(\f)$ becomes zero in $T \sim
O(\log n)$ iterations.

In general, any vector-valued multi-class learning algorithm can be used to
minimize~(\ref{eqnEnergy}). Although this goal is clearly defined
in~\cite{ScSi99}, efficient base learning algorithms have never been described
in detail. In most recent papers \cite{ZhZoRoHa09,MuSc13} where
$\Algo{AdaBoost.MH}$ is used as baseline, the base learner is a classical
single-label decision tree which has to be grown rather large to satisfy the
weak-learning condition, and, when boosted, yields suboptimal results
(Section~\ref{secExperiments}). The reason why methods for learning multi-class
$\{\pm 1\}^K$-valued base classifiers had not been developed before is because
they \emph{have} to be boosted: since they do not select a single label, they
cannot be used as stand-alone multi-class classifiers.


Although it is not described in detail, it seems that the base classifier used
in the original paper of \citet{ScSi99} is a vector of $K$ independent decision
stumps $\bh(\bx) = \big(h_1(\bx),\ldots,h_K(\bx)\big)$. These stumps cannot be
used as node classifiers to grow decision trees since they do not define a
\emph{single} cut that depends only on the input (see
Section~\ref{secHammingTrees} for a more detailed discussion). To overcome this
problem, we propose base learning algorithms that \emph{factorize} $\bh(\bx)$
into
\begin{equation}\label{eqnMulticlassBaseClassifier}
\bh(\bx) = \alpha \bv \varphi(\bx),
\end{equation}
where $\alpha \in \RR^+$ is a positive real valued \emph{base coefficient},
$\bv$ is an input-independent \emph{vote} vector of length $K$, and
$\varphi(\bx)$ is a label-independent \emph{scalar} classifier. In
\emph{discrete} \Algo{AdaBoost.MH}, both components are binary, that is, $\bv
\in \{\pm 1\}^K$ and $\varphi(\bx): \RR^d \rightarrow \{\pm 1\}$. The setup can
be extended to real-valued classifiers $\varphi(\bx): \RR^d \rightarrow \RR$,
also known as \emph{confidence-rated} classifiers, and it is also easy to make
the vote vector $\bv$ real-valued (in which case, without the loss of
generality, $\alpha$ would be set to $1$). Both variants are known under the
name of \emph{real} \Algo{AdaBoost.MH}. Although there might be slight
differences in the practical performance of real and discrete
\Algo{AdaBoost.MH}, here we decided to stick to the discrete case for the sake
of simplicity.

\vspace{\sectionBefore}
\subsection{Casting the votes}
\label{secCastingTheVotes}

To start, we show how to set $\alpha$ and $\bv$ in general if the scalar base
classifier $\varphi$ is given. The intuitive semantics of
(\ref{eqnMulticlassBaseClassifier}) is the following. The binary classifier
$\varphi(\bx)$ cuts the input space into a positive and a negative region. In
binary classification this is the end of the story: we need $\varphi(\bx)$ to be
well-correlated with the binary class labels $y$. In multi-class classification
it is possible that $\varphi(\bx)$ correlates with some of the class labels
$y_\ell$ and anti-correlates with others. This free choice is expressed by the
binary ``votes'' $v_\ell \in \{\pm 1\}$. We say that $\varphi(\bx)$ votes
\emph{for} class $\ell$ if $v_\ell = +1$ and it votes \emph{against} class
$\ell$ if $v_\ell = -1$. As in binary classification, $\alpha$ expresses the
overall quality of the classifier $\bv \varphi(\bx)$: $\alpha$ is monotonically
decreasing with respect to the weighted error of $\bv \varphi(\bx)$.

The advantage of the setup is that, given the binary classifier $\varphi(\bx)$,
the optimal vote vector $\bv$ and the coefficient $\alpha$ can be set in an
efficient way. To see this, first let us define the \emph{weighted per-class
  error rate}
\begin{equation}\label{eqnPerClassErrorRate}
  \mu_{\ell-} = \sum_{i=1}^n w_{i,\ell} \IND{\varphi(\bx_i)
    \not= y_{i,\ell}},
\end{equation}
and the \emph{weighted per-class correct classification rate}
\begin{equation}\label{eqnPerClassCorrectRate}
  \mu_{\ell+} = \sum_{i=1}^n w_{i,\ell} \IND{\varphi(\bx_i) =
    y_{i,\ell}}
\end{equation}
for each class $\ell = 1,\ldots,K$. With this notation, $Z\big(\bh,\bW\big)$
simplifies to (see Appendix~\ref{secFactorizedEnergy})
\begin{equation}\label{eqnFactorizedEnergy}
Z(\bh,\bW) = \frac{e^{\alpha} +e^{-\alpha}}{2} - \frac{e^{\alpha} -
  e^{-\alpha}}{2} \sum_{\ell=1}^K v_\ell \big(\mu_{\ell+} -
\mu_{\ell-}\big).
\end{equation}
The quantity
\begin{equation}\label{eqnClasswiseEdge}
\gamma_\ell = v_\ell \big(\mu_{\ell+} - \mu_{\ell-}\big) = \sum_{i=1}^n
w_{i,\ell} v_\ell \varphi(\bx_i) y_{i,\ell}
\end{equation}
is called the \emph{classwise edge} of $\bh(\bx)$. The full multi-class edge of
the classifier is then
\begin{equation}\label{eqnMulticlassEdge}
\begin{split}
\gamma & = \gamma(\bv,\varphi,\bW) = \sum_{\ell=1}^K \gamma_\ell =
\sum_{\ell=1}^K 
v_\ell \big(\mu_{\ell+} - \mu_{\ell-}\big) \\& = \sum_{i=1}^n \sum_{\ell=1}^K
w_{i,\ell} v_\ell \varphi(\bx_i) y_{i,\ell}.
\end{split}
\end{equation}
With this notation, the classical~\cite{FrSc97} binary coefficient $\alpha$ is
recovered: it is easy to see that (\ref{eqnFactorizedEnergy}) is minimized when
\begin{equation}\label{eqnMulticlassAlpha}
\alpha = \frac{1}{2}\log\frac{1+\gamma}{1-\gamma}.
\end{equation}
With this optimal coefficient, (\ref{eqnFactorizedEnergy}) becomes $Z(\bh,\bW) =
\sqrt{1 - \gamma^2}$, so $Z(\bh,\bW)$ is minimized when $\gamma$ is
maximized. From (\ref{eqnMulticlassEdge}) it then follows that $Z(\bh,\bW)$ is
minimized if $v_\ell$ agrees with the sign of $\big(\mu_{\ell+} -
\mu_{\ell-}\big)$, that is,
\begin{equation}\label{eqnVotes}
  v_\ell = \begin{cases}
    1 & \mbox{ if } \mu_{\ell+} > \mu_{\ell-}\\
    -1 & \mbox{ otherwise}
  \end{cases}
\end{equation}
for all classes $\ell = 1,\ldots,K$. 

The setup of factorized base classification (\ref{eqnMulticlassBaseClassifier})
has another important consequence: the preservation of the weak-learning
condition. Indeed, if $\varphi(\bx)$ is slightly better then a coin toss,
$\gamma$ will be positive. Another way to look at it is to say that if a
$(\varphi,\bv)$ combination has a negative edge $\gamma < 0$, then the edge of
its \emph{complement} (either $(-\varphi,\bv)$ or $(\varphi,-\bv)$) will be
$-\gamma > 0$. To understand the significance of this, consider a classical
single-label base classifier $h: \cX \rightarrow \cL = \{1,\ldots,K\}$, required
by \Algo{AdaBoost.M1}. Now if $h(\bx)$ is slightly better than a coin toss, all
one can hope for is an error rate slightly lower than $\frac{K-1}{K}$ (which is
equivalent to an edge slightly higher than $\frac{2-K}{K}$). To achieve the
error of $\frac{1}{2}$ (zero edge), required for continuing boosting, one has to
come up with a base learner which is significantly better than a coin toss.

There is a long line of research on \emph{output codes} similar in spirit to our
setup. The boosting engine in these works is usually slightly different from
\Algo{AdaBoost.MH} since it attempts to optimize the multi-class hinge loss, but
the factorization of the multi-class base classifier is similar to
(\ref{eqnMulticlassBaseClassifier}). Formally, the vote vector $\bv$ in this
framework is one column in an output code matrix. In the simplest setup this
matrix is fixed beforehand by maximizing the error correcting capacity of the
matrix \cite{DiBa95, AlScSi01}. A slightly better solution
\cite{Sch97,GuSa99,SuToLiWu05} is to wait until the given iteration to pick
$\bv$ by maximizing
\[
\bv^* = \argmax_\bv \sum_{i=1}^n \sum_{\ell=1}^K w_{i,\ell} \IND{v_\ell \not=
  v_{\ell(\bx_i)}},
\]
and then to choose the optimal binary classifier $\varphi$ with this fixed vote
(or code) vector $\bv^*$ (although in practice it seems to be better to fix
$\bv$ to a random binary vector; \citealt{SuToLiWu05}). The state of the art in
this line of research is to iterate between optimizing $\varphi$ with a fixed
$\bv$ and then picking the best $\bv$ with a fixed $\varphi$ \cite{Li06,KeBu09,
  GaKo11a}.

It turns out that if $\varphi$ is a decision stump, exhaustive search for
\emph{both} the best binary cut (threshold) and the best vote vector can be
carried out using one single sweep in $\Theta(nK)$ time. The algorithm is a
simple extension of the classical binary decision stump learner; for the sake of
completeness, we provide the pseudocode in Appendix~\ref{secStumpBase}. The
computational efficiency of this learning algorithm combined with the factorized
form (\ref{eqnMulticlassBaseClassifier}) of the classifier allows us to build
multiclass Hamming trees in an efficient manner, circumventing the problem of
global maximization of the edge with respect to $\varphi$ and $\bv$.

\vspace{\sectionBefore}
\section{Hamming trees}
\vspace{\sectionAfter}
\label{secHammingTrees}

Classification trees \cite{Qui86} have been widely used for multivariate
classification since the 80s. They are especially efficient when used as base
learners in \Algo{AdaBoost}~\cite{CaNi06,Qui96}. Their main disadvantage is
their variance with respect to the training data, but when averaged over $T$
different runs, this problem largely disappears. The most commonly used tree
learner is \Algo{C4.5} of \citet{Qui93}. Whereas this tree implementation is a
perfect choice for binary \Algo{AdaBoost}, it is suboptimal for
\Algo{AdaBoost.MH} since it outputs a single-label classifier with no guarantee
of a positive multi-class edge (\ref{eqnMulticlassEdge}). Although this problem
can be solved in practice by building large trees, it seems that using these
large single-class trees is suboptimal (Section~\ref{secExperiments}).

The main technical difficulty of building trees out of generic $\{\pm
1\}^K$-valued multi-class classifiers $\bh(\bx)$ is that they do not necessarily
implement a binary cut $\bx \mapsto \{\pm 1\}$, and partitioning the data into
all the possibly $2^K$ children at a tree node leads to rapid
overfitting. Factorizing the multi-class classifier $\bh(\bx)$ into an
input-independent vote vector $\bv$ and a label-independent binary classifier
$\varphi(x)$ as in (\ref{eqnMulticlassBaseClassifier}) solves this problem. Base
classifiers are trained as usual at each new tree leaf. In case this leaf
remains a leaf, the full classifier $\bh(\bx)$ is used for instances $\bx$ that
arrive to this leaf. If it becomes an inner node, the vote vector $\bv$ is
discarded, and the partitioning of the data set is based on solely the binary
classifier $\varphi(x)$. An advantage of this formalization is that we can use
any multi-class base classifier of the form (\ref{eqnMulticlassBaseClassifier})
for the tree cuts, so the Hamming tree algorithm can be considered as a ``meta
learner'' which can be used on the top of any factorized base learner.

Formally, a binary classification tree with $N$ inner nodes ($N+1$ leaves)
consists of a list of $N$ base classifiers $\bgH = (\bh_1,\ldots,\bh_N)$ of the
form $\bh_j(\bx) = \alpha_j \bv_j \varphi_j(\bx)$ and two index lists $\bgl =
(\gl_1,\ldots,\gl_N)$ and $\bgr = (\gr_1,\ldots,\gr_N)$ with $\bgl,\bgr \in (\N
\cup \{\Algo{null}\})^N$. $\gl_j$ and $\gr_j$ represent the indices of the left
and right children of the $j$\/th node of the tree, respectively. The
\emph{node} classifier in the $j$\/th node is defined recursively as
\begin{equation}\label{eqnTreeClassifier}
\bgh_j(\bx) = \begin{cases}
-\bv_{j} & \text{if } \varphi_j(\bx) = -1 \wedge \gl_j = \Algo{null}\\& \text{\;\;
  (left leaf),} \\  
\bv_{j} & \text{if } \varphi_j(\bx) = +1 \wedge \gr_j = \Algo{null}\\& \text{\;\;
  (right leaf),} \\  
\bgh_{\gl_j}(\bx) & \text{if } \varphi_j(\bx) = -1 \wedge \gl_j \not=
\Algo{null}\\& \text{\;\;  (left inner node),}\\ 
\bgh_{\gr_j}(\bx) & \text{if } \varphi_j(\bx) = +1 \wedge \gr_j \not=
\Algo{null}\\& \text{\;\;  (right inner node).}
\end{cases}
\end{equation}
The final tree classifier $\bh_{\bgH,\bgl,\bgr}(\bx) = \alpha \bgh_1(\bx)$
itself is not a factorized classifier
(\ref{eqnMulticlassBaseClassifier}).\footnote{Which is not a problem: we will
  not want to build trees of trees.} In particular, $\bh_{\bgH,\bgl,\bgr}(\bx)$
uses the local vote vectors $\bv_j$ determined by each leaf instead of a global
vote vector. On the other hand, the coefficient $\alpha$ is unique, and it is
determined in the standard way
\[
\alpha = \frac{1}{2} \log \frac{1 + \gamma(\bgh_1,\bW)}{1 -
      \gamma(\bgh_1,\bW)}
\]
based on the edge of the tree classifier $\bgh_1$. The local coefficients
$\alpha_j$ returned by the base learners are discarded (along with the vote
vectors in the inner nodes).

Finding the optimal $N$-inner-node tree is a difficult combinatorial
problem. Most tree-building algorithms are therefore sub-optimal by
construction. For \Algo{AdaBoost} this is not a problem: we can continue
boosting as long as the edge is positive. Classification trees are usually built
in a greedy manner: at each stage we try to cut all the current leaves $j$ by
calling the base learner of the data points reaching the $j$\/th leaf, then
select the best node to cut, convert the old leaf into an inner node, and add
two new leaves. The difference between the different algorithms is in the way
the best node is selected. Usually, we select the node that \emph{improves} a
gain function the most. In \Algo{AdaBoost.MH} the natural gain is the edge
(\ref{eqnMulticlassEdge}) of the base classifier. Since the data set $(\bX,\bY)$
is different at each node, we include it explicitly in the argument of the full
multi-class edge
\[
\gamma(\bv,\varphi,\bX,\bY,\bW) = \sum_{i=1}^n \sum_{\ell=1}^K \IND{x_i
  \in \bX} w_{i,\ell} v_\ell \varphi(\bx_i) y_{i,\ell}.
\]
Note that in this definition we do not require that the weights of the selected
points add up to $1$. Also note that this gain function is additive on subsets
of the original data set, so the local edges in the leaves add up to the edge of
the full tree. This means that any improvement in the local edge directly
translates to an improvement of the tree edge. This is a crucial property: it
assures that the edge of the tree is always positive as long as the local edges
in the inner nodes are positive, so any weak binary classifier $\phi(\bx)$ can
be used to define the inner cuts and the leaves.

The basic operation when adding a tree node with a scalar binary classifier
(cut) $\varphi$ is to separate the data matrices $\bX$, $\bY$, and $\bW$
according to the sign of classification $\varphi(\bx_i)$ for all $\bx_i \in
\bX$. The pseudocode is straightforward, but for the sake of completeness, we
include it in the supplementary (Appendix~\ref{secCutDataSetProductBase},
Figure~\ref{figCutDataSetProductBase}).

Building a tree is usually described in a recursive way but we find the
iterative procedure easier to explain, so our pseudocode in
Figure~\ref{figTreeBase} contains this version. The main idea is to maintain a
\emph{priority queue}, a data structure that allows \emph{inserting} objects
with numerical \emph{keys} into a set, and extracting the object with the
\emph{maximum} key \cite{CoLeRi94}. The key will represent the improvement of
the edge when cutting a leaf. We first call the base learner on the full data
set (line~\ref{linTreeFirstBase}) and insert it into the priority queue with its
edge $\gamma(\bv,\varphi,\bX,\bY,\bW)$ (line~\ref{linTreeFirstInsertion}) as the
key. Then in each iteration, we extract the leaf that would provide the best
edge improvement among all the leaves in the priority queue
(line~\ref{linTreeBestExctraction}), we partition the data set
(line~\ref{linTreeCut}), call the base learners on the two new leaves
(line~\ref{linTreeLeafBase}), and insert them into the priority queue using the
difference between the old edge \emph{on the partitioned data sets} and the new
edges of the base classifiers in the two new leaves
(line~\ref{linTreeLeafInsertion}). When inserting a leaf into the queue, we also
save the sign of the cut (left or right child) and the index of the parent, so
the index vectors $\bgl$ and $\bgr$ can be set properly in
line~\ref{linTreeSetIndexVectors}.

\setlength{\algowidth}{\textwidth}
\begin{figure*}[!ht]
\centerline{
  \begin{algorithm}{$\Algo{TreeBase}(\bX,\bY,\bW,\Algo{Base}(\cdot,\cdot,\cdot),N)$}
    \Aitem $\big(\alpha,\bv,\varphi(\cdot)\big) \setto
    \Algo{Base}(\bX,\bY,\bW)$ \label{linTreeFirstBase}
    \Aitem $S \setto \Algo{PriorityQueue}$ \mt \algoremark{$O(\log N)$ insertion
      and extraction of maximum key}
    \Aitem $\Algo{Insert}\big(S,
    \big(\bv,\varphi(\cdot),\bX,\bY,\Algo{null},0\big), 
    \gamma(\bv,\varphi,\bX,\bY,\bW)\big)$ \mt \algoremark{key $=$ edge
      $\gamma$} \label{linTreeFirstInsertion} 
    \Aitem $\bgH \setto ()$  \mt
    \algoremark{initialize classifier list}
    \Aitem \For $j \setto 1$ \To $N$
    \Aitem \mt $\gl_j \setto \gr_j \setto \Algo{null}$  \mt
    \algoremark{initialize child indices}
    \Aitem \mt
    $\big(\bv_j, \varphi_j(\cdot), \bX_j, \bY_j, \bullet,
    j_\Algo{p}\big) \setto \Algo{ExtractMax}(S)$ \mt \algoremark{best node in
      the priority queue} \label{linTreeBestExctraction}
    \Aitem \mt \If $\bullet = -$ \Then $\gl_{j_\Algo{p}} \setto j$ \Else
    \If $\bullet = +$ \Then $\gr_{j_\Algo{p}} \setto j$ \mt \algoremark{child
      index of parent} \label{linTreeSetIndexVectors}
    \Aitem \mt $\bgH \setto \Algo{Append}(\bgH,\bv_j \varphi_j(\cdot))$ \mt
    \algoremark{adding $\bh_j(\cdot) = \bv_j \varphi_j(\cdot)$ to $\bgH$}
    \Aitem \mt $(\bX_-, \bY_-, \bW_-, \bX_+, \bY_+, \bW_+) \setto
    \Algo{CutDataSet}\big(\bX_j,\bY_j,\bW,\varphi_j(\cdot)\big)$ 
    \Aitem \mt \For $\bullet \in \{-,+\}$ \mt \algoremark{insert children
      into priority queue} \label{linTreeCut}
    \Aitem \mtt $\big(\alpha_\bullet,\bv_\bullet,\varphi_\bullet(\cdot)\big) \setto
    \Algo{Base}(\bX_\bullet,\bY_\bullet,\bW_\bullet)$ \label{linTreeLeafBase}
    \Aitem    \label{linTreeLeafInsertion} \mtt $\!\!\Algo{Insert}\big(S,
    \big(\bv_\bullet,\varphi_\bullet(\cdot),\bX_\bullet,\bY_\bullet, \bullet,
    j\big), 
    \gamma(\bv_\bullet,\varphi_\bullet,\!\bX_\bullet,\!\!\bY_\bullet,\!\!\bW_\bullet) -
    \gamma(\bv_j,\varphi_j,\!\bX_\bullet,\!\!\bY_\bullet,\!\!\bW_\bullet)\big)$ 
    \AitemNoLabel \mttt \algoremark{key $= $ edge improvement over parent edge}
    \Aitem $\displaystyle \alpha = \frac{1}{2} \log \frac{1 + \gamma(\bgh_1,\bW)}{1 -
      \gamma(\bgh_1,\bW)}$  \mt \algoremark{standard coefficient of the full
      tree classifier $\bgh_1$ (\ref{eqnTreeClassifier})}
    \Aitem \Return $\big(\alpha, \bgH, \bgl, \bgr\big)$ 
  \end{algorithm}
}
  \caption{The pseudocode of the Hamming tree base learner. $N$ is the number of
    inner nodes. The algorithm returns a list of base classifiers $\bgH$, two
    index lists $\bgl$ and $\bgr$, and the base coefficient $\alpha$. The tree
    classifier is then defined by
    (\ref{eqnTreeClassifier}). \label{figTreeBase}}
\vspace{-0.0cm}
\end{figure*}

When the priority queue is implemented as a heap, both the insertion and the
extraction of the maximum takes $O(\log N)$ time~\cite{CoLeRi94}, so the total
running time of the procedure is $O\big(N(T_\Algo{Base} + n + \log N)\big)$,
where $T_\Algo{Base}$ is the running time of the base learner. Since $N$ cannot
be more than $n$, the running time is $O\big(N(T_\Algo{Base} + n)\big)$. If the
base learners cutting the leaves are decision stumps, the total running time is
$O(nKdN)$. In the procedure we have no explicit control over the shape of the
tree, but if it happens to be balanced, the running time can further be improved
to $O(nKd\log N)$.

\vspace{0cm}
\section{Experiments}
\vspace{0cm}
\label{secExperiments}

Full reproducibility was one of the key motivations when we designed our
experimental setup. All experiments were done using the open source
\href{http://multiboost.org}{multiboost software} of \citet{BBCCK11},
version~1.2. In addition, we will make public all the configuration files,
train/test/validation cuts, and the scripts that we used to set up the
hyperparameter validation.

We carried out experiments on five mid-sized (isolet, letter, optdigits,
pendigits, and USPS) and nine small (balance, blood, wdbc, breast, ecoli, iris,
pima, sonar, and wine) data sets from the
\href{http://archive.ics.uci.edu/ml/datasets.html}{UCI repository}. The five
sets were chosen to overlap with the selections of most of the recent
multi-class boosting papers~\cite{KeBu09,Li09,Li09a,ZhZoRoHa09,SuReZh12,MuSc13},
The small data sets were selected for comparing \Algo{AdaBoost.MH} with SVMs
using Gaussian kernels, taking the results of a recent paper~\cite{DuJaMa12}
whose experimental setup we adopted. All numerical results (multi-class test
errors $\widehat{R}_\I(\f)$ (\ref{eqnOneError}) and test learning curves) are
available at \url{https://www.lri.fr/~kegl/research/multiboostResults.pdf}, one
experiment per page for clarity.  Tables~\ref{tabResults}
and~\ref{tabResultsSmall} contain summaries of the results.

\begin{table*}[!ht]
\small
\centering
\begin{tabular}{|l|c|c|c|c|c|}
\hline
\textbf{Method} & \textbf{isolet} & \textbf{letter} &  \textbf{optdigits} & \textbf{pendigits} & \textbf{USPS}\\
\hline
\Algo{AdaBoost.MH} w Hamming trees & $3.5 \pm 0.5$ & $2.1 \pm 0.2$ & $2.0 \pm 0.3$ & $2.1 \pm 0.3$ & $4.5 \pm 0.5$\\
\hline
\Algo{AdaBoost.MH} w Hamming prod. \cite{KeBu09} & $4.2 \pm 0.5$ & $2.5 \pm 0.2$
& $2.1 \pm 0.4$ & $2.1 \pm 0.2$ & $4.4 \pm 0.5$\\
\hline
\Algo{AOSOLogitBoost} $J = 20$, $\nu = 0.1$ \cite{SuReZh12} & $3.5 \pm 0.5$ & $2.3 \pm 0.2$ & $2.1 \pm 0.3$ & $2.4 \pm 0.3$ & $4.9 \pm 0.5$\\
\hline
\Algo{ABCLogitBoost} $J = 20$, $\nu = 0.1$ \cite{Li09a} & $4.2 \pm 0.5$ & $2.2 \pm 0.2$ & $3.1 \pm 0.4$ & $2.9 \pm 0.3$ & $4.9 \pm 0.5$\\
\hline
\Algo{ABCMart} $J = 20$, $\nu = 0.1$ \cite{Li09} & $5.0 \pm 0.6$ & $2.5 \pm 0.2$ & $2.6 \pm 0.4$ & $3.0 \pm 0.3$ & $5.2 \pm 0.5$\\
\hline
\Algo{LogitBoost} $J = 20$, $\nu = 0.1$ \cite{Li09a} & $4.7 \pm 0.5$ & $2.8 \pm 0.3$ & $3.6 \pm 0.4$ & $3.1 \pm 0.3$ & $5.8 \pm 0.5$\\
\hline
\Algo{SAMME} w single-label trees \cite{ZhZoRoHa09} & & $2.3 \pm 0.2$ & & $2.5 \pm 0.3$ & \\
\hline
\Algo{AdaBoost.MH} w single-label trees \cite{ZhZoRoHa09} & & $2.6 \pm 0.3$ &
& $2.8 \pm 0.3$ & \\
\hline
\Algo{AdaBoost.MM} \cite{MuSc13} & & $2.5 \pm 0.2$ & & $2.7 \pm 0.3$ & \\
\hline
\Algo{AdaBoost.MH} w single-label trees \cite{MuSc13} & & $9.0 \pm 0.5$ & &
$7.0 \pm 0.4$ & \\
\hline
\end{tabular}
\caption{Test error percentages on mid-sized benchmark data
  sets.\label{tabResults}}
\end{table*}

\begin{table}[!ht]
\small
\centering
\begin{tabular}{|l|c|c|}
\hline
& \textbf{\Algo{AB.MH}} & \textbf{SVM}\\
\hline
balance & $6.0 \pm 4.0$ & $10.0 \pm 2.0$\\
blood & $22.0 \pm 4.0$ & $21.0 \pm 5.0$\\
wdbc & $3.0 \pm 2.0$ & $2.0 \pm 3.0$\\
breast & $34.0 \pm 13.0$ & $37.0 \pm 8.0$\\
ecoli & $15.0 \pm 6.0$ & $16.0 \pm 6.0$\\
iris & $7.0 \pm 6.0$ & $5.0 \pm 6.0$\\
pima & $24.0 \pm 5.0$ & $24.0 \pm 4.0$\\
sonar & $13.0 \pm 10.0$ & $14.0 \pm 8.0$\\
wine & $2.0 \pm 3.0$ & $3.0 \pm 4.0$\\
\hline
\end{tabular}
\caption{Test error percentages on small benchmark data
  sets.\label{tabResultsSmall}}
\end{table}

Hyperparameter optimization is largely swept under the rug in papers describing
alternative multi-class boosting methods. Some report results with fixed
hyperparameters \cite{ZhZoRoHa09,SuReZh12} and others give the full table of
test errors for a grid of hyperparameters \cite{KeBu09,Li09,Li09a,MuSc13}.
Although the following procedure is rather old, we feel the need to detail it
for promoting a more scrupulous comparison across papers.

For the small data sets we ran $10\times 10$ cross-validation (CV) to optimize
the hyperparameters and the estimate the generalization error. For the number of
inner nodes we do a grid search (we also considered using the ``one sigma'' rule
for biasing the selection towards smaller trees, but the simple minimization
proved to be better). For robustly estimating the optimal stopping time we use a
smoothed test error. For the formal description, let $\widehat{R}^{(t)}$ be the
average test error (\ref{eqnOneError}) of the ten validation runs after $t$
iterations. We run \Algo{AdaBoost.MH} for $T_{\max}$ iterations, and compute the
optimal stopping time using the minimum of the smoothed test error using a
linearly growing sliding window, that is,
\begin{equation}\label{eqnStoppingTime}
T^* = \argmin_{T:T_{\min} < T \le T_{\max}}\frac{1}{T - \lfloor 0.8T
  \rfloor}\sum_{t=\lfloor 0.8T \rfloor}^T \widehat{R}^{(t)},
\end{equation}
where $T_{\min}$ was set to a constant $50$ to avoid stopping too early due to
fluctuations. 
For selecting the best number of inner nodes $N$, we simply minimized the
smoothed test error over a predefined grid
\[
N^* = \min_{N \in \cN} \widehat{R}^{(T^*_N)}(N)
\]
where $T^*_{N}$ and $\widehat{R}^{(t)}(N)$ are the optimal stopping time
(\ref{eqnStoppingTime}) and the test error, respectively, in the run with $N$
inner nodes, and $\cN$ is the set of inner nodes participating in the grid
search.
Then we re-run \Algo{AdaBoost.MH} on the joined training/validation set using
the selected hyperparameters $N^*$ and $T^*_{N^*}$. The error $\widehat{R}_i$ in
the $i$\/th training/test fold is then computed on the held-out test set. In the
tables we report the mean error and the standard deviation. On the medium-size
data sets we ran $1\times 5$ CV (using the designated test sets where available)
following the same procedure. In this case the report the binomial standard
deviation $\sqrt{\widehat{R}(1-\widehat{R})/n}$. Further details and the
description and explanation of some slight variations of this experimental setup
are available at \url{https://www.lri.fr/~kegl/research/multiboostResults.pdf}.

On the small data sets, \citet{DuJaMa12} used the exact same protocol, so,
although the folds are not the same, the results are directly comparable. The
error bars represent the standard deviation of the test errors over the ten test
folds \emph{not} divided by $\sqrt{10}$, contrary to common practice, since the
training set of the folds are highly correlated. The large error bars are the
consequence of the small size and the noisiness of these sets. They make it
difficult to establish any significant trends. We can safely state that
\Algo{AdaBoost.MH} is on par with SVM (it is certainly not worse, ``winning'' on
six of the nine sets), widely considered one of the the best classification
methods for small data sets.

Even though on the mid-sized data sets there are dedicated test sets used by
most of the experimenters, comparing \Algo{AdaBoost.MH} to alternative
multi-class boosting techniques is somewhat more difficult since none of the
papers do proper hyperparameter tuning. Most of the papers report results with a
table of errors given for a set of hyperparameter choices, without specifying
which hyperparameter choice would be picked by proper validation. For methods
that are non-competitive with \Algo{AdaBoost.MH} (\Algo{SAMME} of
\citet{ZhZoRoHa09} and \Algo{AdaBoost.MM} of \citet{MuSc13}) we report the
\emph{post-validated} best error which may be significantly lower than the error
corresponding to the hyperparameter choice selected by proper validation. For
methods where this choice would unfairly bias the comparison
(\Algo{AOSOLogitBoost}~\cite{SuReZh12}, \Algo{ABCLogitBoost}, \Algo{LogitBoost},
and \Algo{ABCMart}~\cite{Li09,Li09a}), we chose the best overall hyperparameter
$J = 20$ and $\nu = 0.1$, suggested by the \citet{Li09,Li09a}. At
\url{https://www.lri.fr/~kegl/research/multiboostResults.pdf} (but not in
Table~\ref{tabResults}) we give both errors for some of the methods. Proper
hyperparameter-validation should put the correct test error estimates between
those two limits. Since \Algo{AdaBoost.MH} with decision products~\cite{KeBu09}
is also implemented in \href{http://multiboost.org}{multiboost}~\cite{BBCCK11},
for this method we re-ran experiments with the protocol described above.


The overall conclusion is that \Algo{AOSOLogitBoost}~\cite{SuReZh12} and
\Algo{AdaBoost.MH} with Hamming trees are the best algorithms
(\Algo{AdaBoost.MH} winning on all the five data sets but within one standard
deviation). \Algo{AdaBoost.MH} with decision products~\cite{KeBu09} and
\Algo{ABCLogitBoost} are slightly weaker, as also noted
by~\cite{SuReZh12}. \Algo{SAMME}~\cite{ZhZoRoHa09} and
\Algo{AdaBoost.MM}~\cite{MuSc13} perform below the rest of the methods on the
two data sets shared among all the papers (even though we give post-validated
results). Another important conclusion is that \Algo{AdaBoost.MH} with Hamming
trees is significantly better then other implementations of \Algo{AdaBoost.MH}
in~\cite{ZhZoRoHa09,MuSc13}, assumably implemented using single-label trees (the
errors reported by \citet{MuSc13} are especially conspicuous).

\Algo{AdaBoost.MH} with Hamming trees also achieves good results on image
recognition problems. On \href{http://yann.lecun.com/exdb/mnist}{MNIST},
boosting trees of stumps over pixels with eight inner nodes and about $50000$
iterations has a test error of $1.25\%$, making it one of the best
no-domain-knowledge ``shallow'' classifiers. Using stumps over Haar filters
\cite{ViJo01}, boosted trees with four inner nodes and $10000$ iterations
achieves a test error of $0.85\%$, comparable to classical convolutional nets
\cite{LeBoBeHa98}.

\Algo{AdaBoost.MH} with Hamming trees, usually combined with calibration
\cite{Pla99,NiCa05} and model averaging, has been also successful in recent data
challenges. On the
\href{http://www.kaggle.com/c/challenges-in-representation-learning-facial-expression-recognition-challenge/leaderboard/private}{Kaggle
  emotions} data challenge, although not competitive with deep learning
techniques, out-of-the-box \Algo{AdaBoost.MH} with Hamming trees over Haar
filters finished $17$\/th place with a test error of $57\%$. In the Yahoo!
Learning-to-Rank Challenge \cite{ChChLi11a} it achieved top ten performances
with results not significantly different from the winning scores. Finally, in
the recent
\href{http://emotion-research.net/sigs/speech-sig/is13-compare}{INTERSPEECH
  Challenge} it won the Emotion sub-challenge and it was runner up in the Social
Signals sub-challenge.


\vspace{\sectionBefore}
\section{Conclusion}
\vspace{\sectionAfter}
\label{secConclusion}

In this paper we introduced Hamming trees that optimize the multi-class edge
prescribed by \Algo{AdaBoost.MH} without reducing the multi-class problem to $K$
binary one-against-all classifications. We showed that without this restriction,
often considered mandatory, \Algo{AdaBoost.MH} is one of the best off-the-shelf
multi-class classification algorithms. The algorithm retains the conceptual
elegance, power, and computational efficiency of binary \Algo{AdaBoost}.

Using decision stumps at the inner nodes and at the leaves of the tree is a
natural choice due to the efficiency of the learning algorithm, nevertheless,
the general setup described in this paper allows for using \emph{any} binary
classifier. One of the avenues investigated for future work is to try stronger
classifiers, such as SVMs, as binary cuts. The formal setup described in
Section~\ref{secBoostingPreliminaries} does not restrict the algorithm to
single-label problems; another direction for future work is to benchmark it on
standard multi-label and sequence-to-sequence classification problems
\cite{DiHaAs08}.

\newpage
\onecolumn

\appendix

\section{Showing (\ref{eqnFactorizedEnergy})}
\label{secFactorizedEnergy}

\begin{eqnarray}
Z(\bh,\bW) & = & \sum_{i=1}^n \sum_{\ell=1}^K w_{i,\ell}
\exp\big(\!\!-\!h_\ell(\bx_i) y_{i,\ell}\big) = \sum_{i=1}^n \sum_{\ell=1}^K
w_{i,\ell} \exp\big(\!\!-\!\alpha v_\ell \varphi(\bx_i)
y_{i,\ell}\big)\label{eqnMulticlassBaseClassifierUsed}\\ 
& = &
\sum_{i=1}^n \sum_{\ell=1}^K \Big(w_{i,\ell} \IND{v_\ell \varphi(\bx_i)
  y_{i,\ell} = 1}e^{-\alpha} + w_{i,\ell}
\IND{v_\ell \varphi(\bx_i) y_{i,\ell} = -1} e^{\alpha}\Big)\nonumber\\ 
& = &
\sum_{\ell=1}^K \big(\mu_{\ell+} \IND{v_\ell = +1}
  + \mu_{\ell-} \IND{v_\ell =
    -1}\big)e^{-\alpha}\nonumber\\
&& + \sum_{\ell=1}^K \big(\mu_{\ell-} \IND{v_\ell = +1} + \mu_{\ell+}
\IND{v_\ell = -1}\big)e^{\alpha} \label{eqnPerClassErrorRateUsed}\\ 
& = &
\sum_{\ell=1}^K \Big(\IND{v_\ell = +1}\big(e^{-\alpha}\mu_{\ell+}
  + e^{\alpha} \mu_{\ell-}\big) + \IND{v_\ell = -1}\big(e^{-\alpha}\mu_{\ell-} 
  + e^{\alpha} \mu_{\ell+}\big)\Big) \nonumber\\ 
& = &
\sum_{\ell=1}^K \left(\frac{1+v_\ell}{2}\big(e^{-\alpha}\mu_{\ell+} 
  + e^{\alpha} \mu_{\ell-}\big) + \frac{1-v_\ell}{2}\big(e^{-\alpha}\mu_{\ell-}
  + e^{\alpha} \mu_{\ell+}\big)\right) \nonumber\\ 
& = &
\frac{1}{2}\sum_{\ell=1}^K \Big(\big(e^{\alpha} +e^{-\alpha}\big) \big(\mu_{\ell+} +
\mu_{\ell-} \big) - v_\ell \big(e^{\alpha} - e^{-\alpha}\big)\big(\mu_{\ell+} -
\mu_{\ell-}\big)\Big)\nonumber\\ 
& = &
\frac{e^{\alpha} +e^{-\alpha}}{2} - \frac{e^{\alpha} - e^{-\alpha}}{2} \sum_{\ell=1}^K
v_\ell \big(\mu_{\ell+} -
\mu_{\ell-}\big).\label{eqnNormalizedWeightUsed}
\end{eqnarray}
(\ref{eqnMulticlassBaseClassifierUsed}) comes from the definition
(\ref{eqnMulticlassBaseClassifier}) of $\bh$ and
(\ref{eqnPerClassErrorRateUsed}) follows from the definitions
(\ref{eqnPerClassErrorRate}) and (\ref{eqnPerClassCorrectRate}) of $\mu_{\ell-}$
and $\mu_{\ell+}$. In the final step (\ref{eqnNormalizedWeightUsed}) we used the
fact that 
\[
\sum_{\ell=1}^K \big(\mu_{\ell+} + \mu_{\ell-}\big) = \sum_{i=1}^n
\sum_{\ell=1}^K w_{i,\ell} = 1.
\]

\section{Multi-class decision stumps}
\label{secStumpBase}

The simplest scalar base learner used in practice on numerical features is the
  \emph{decision stump}, a one-decision two-leaf decision tree of the form
\[
\varphi_{j,b}(\bx) = \begin{cases}
  \phantom{-}1 & \mbox{ if } x^{(j)} \geq b, \\
  -1 & \mbox{ otherwise,}
\end{cases}
\]
where $j$ is the index of the selected feature and $b$ is the decision
threshold. If the feature values $\big(x_1^{(j)},\ldots,x_n^{(j)}\big)$ are
pre-ordered before the first boosting iteration, a decision stump maximizing the
edge (\ref{eqnMulticlassEdge}) (or minimizing the energy
(\ref{eqnMulticlassBaseClassifierUsed})\footnote{Note the distinction: for full
  binary $\bv$ the two are equivalent, but for ternary or real valued $\bv$
  and/or real valued $\phi(\bx)$ they are not. In Figure~\ref{figStumpBase} we
  are maximizing the edge within each feature (line~\ref{linStumpFound} in
  \Algo{BestStump}) but across features we are minimizing the energy
  (line~\ref{linMinimumEnergy} in \Algo{StumpBase}). Updating the energy inside
  the inner loop (line~\ref{linStumpInnerLoop}) could not be done in $\Theta(K)$
  time.}) can be found very efficiently in $\Theta(ndK)$ time.

The pseudocode of the algorithm is given in
Figure~\ref{figStumpBase}. \Algo{StumpBase} first calculates the edge vector
$\bgamma^{(0)}$ of the constant classifier $\bh^{(0)}(\bx) \equiv \bOne$ which
will serve as the initial edge vector for each featurewise edge-maximizer. Then
it loops over the features, calls \Algo{BestStump} to return the best
featurewise stump, and then selects the best of the best by minimizing the
energy (\ref{eqnMulticlassBaseClassifierUsed}). \Algo{BestStump} loops over all
(sorted) feature values $s_1,\ldots,s_{n-1}$. It considers all thresholds $b$
halfway between two non-identical feature values $s_i \not = s_{i+1}$. The main
trick (and, at the same time, the bottleneck of the algorithm) is the update of
the classwise edges in lines~\ref{linStumpInnerLoop}-\ref{linStumpUpdate}: when
the threshold moves from $b =\frac{s_{i-1} + s_{i}}{2}$ to $b =\frac{s_{i} +
  s_{i+1}}{2}$, the classwise edge $\gamma_\ell$ of $\bOne \varphi(\bx)$ (that
is, $\bv \varphi(\bx)$ with $\bv = \bOne$) can only change by $\pm w_{i,\ell}$,
depending on the sign $y_{i,\ell}$ (Figure~\ref{figStumpUpdate}). The total edge
of $\bv \varphi(\bx)$ with optimal votes (\ref{eqnVotes}) is then the
sum of the absolute values of the classwise edges of $\bOne \varphi(\bx)$
(line~\ref{linStumpFound}).

\begin{figure}[!ht]
  \centerline{
    \begin{algorithm}{$\Algo{StumpBase}(\bX,\bY,\bW)$}
      \Aitem \For $\ell \setto 1$ \To $K$  \mt \algoremark{for all classes} 
      \Aitem \mt $\displaystyle \gamma^{(0)}_\ell \setto \displaystyle \sum_{i=1}^n
      w_{i,\ell} y_{i,\ell}$ \mt 
      \algoremark{classwise edges (\ref{eqnClasswiseEdge}) of constant
        classifier 
        $\bh^{(0)}(\bx) \equiv \bOne$} 
      \Aitem \For $j \setto 1$ \To $d$  \mtt \algoremark{all (numerical)
        features} 
      \Aitem \mt $\bs \setto \Algo{Sort}\big(x_1^{(j)},\ldots,x_n^{(j)}\big)$
      \mt \algoremark{sort the $j$\/th column of $\bX$} \label{linStumpSort}
      \Aitem \mt $\displaystyle
      (\bv_j,b_j,\gamma_j) \setto
      \Algo{BestStump}(\bs,\bY,\bW,\bgamma^{(0)})$ \mt \algoremark{best stump per
        feature} 
      \Aitem \mt $\displaystyle \alpha_j \setto \frac{1}{2} \log \frac{1 +
        \gamma_j}{1 -
        \gamma_j}$ \mt \algoremark{base coefficient
        (\ref{eqnMulticlassAlpha})} 
      \Aitem $\displaystyle j^* \setto \argmin_j
      Z\big(\alpha_j\bv_j\varphi_{j,b_j},\bW \big)$ \mt \algoremark{best stump
        across features} \label{linMinimumEnergy}
      \Aitem \Return
      $\big(\alpha_{j^*},\bv_{j^*},\varphi_{j^*,{b_{j^*}}}(\cdot)\big)$ 
    \end{algorithm}
  }
  \centerline{
    \begin{algorithm}{$\Algo{BestStump}(\bs,\bY,\bW,\bgamma^{(0)})$} 
    \Aitem $\displaystyle \bgamma^* \setto \bgamma^{(0)}$ \mt
      \algoremark{best edge vector}
    \Aitem $\displaystyle \bgamma \setto \bgamma^{(0)}$ \mt \algoremark{initial
      edge vector} \label{linStumpInitialization}
    \Aitem \For $i \setto 1$ \To $n-1$  \mt \algoremark{for all points in
      order $s_1 \le \ldots \le s_{n-1}$}
    \Aitem \mt \For $\ell \setto 1$ \To $K$  \mt \algoremark{for all
      classes} \label{linStumpInnerLoop} 
    \Aitem \mtt $\displaystyle \gamma_\ell \setto \gamma_\ell - 2w_{i,\ell}
    y_{i,\ell}$ \mt
    \algoremark{update classwise edges of stump with $\bv =
      \bOne$} \label{linStumpUpdate} 
    \Aitem \mt \If $s_{i} \not= s_{i+1}$ \Then \mt \algoremark{no
    threshold if identical coordinates $s_{i} =
    s_{i+1}$} \label{linStumpNonIdentical} 
    \Aitem \mtt \If $\sum_{\ell=1}^K |\gamma_\ell| >
    \sum_{\ell=1}^K |\gamma_\ell^*|$ \Then \mt \algoremark{found better stump}
    \label{linStumpFound} 
    \Aitem \mttt $\bgamma^* \setto \bgamma$ \mt \algoremark{update best edge
      vector} 
    \Aitem \mttt $b^* \setto \frac{s_{i} + s_{i+1}}{2}$ \mt
    \algoremark{update best threshold} 
    \Aitem \For $\ell \setto 1$ \To $K$  \mt \algoremark{for all classes} 
    \Aitem \mt $v^*_\ell \setto \sign(\gamma_\ell)$\mt \algoremark{set vote vector
      according to (\ref{eqnVotes})} 
    \Aitem \If $\bgamma^* = \bgamma^{(0)}$ \mtt \algoremark{did not beat the
      constant classifier} \label{linStumpReturnConstant}
    \Aitem \mt \Return $(\bv^*,-\infty,\|\bgamma^*\|_1)$ \mt \algoremark{
    constant classifier with optimal votes}
    \Aitem \Else
    \Aitem \mt \Return $(\bv^*,b^*,\|\bgamma^*\|_1)$  \mt
    \algoremark{best stump} 
  \end{algorithm}
  }
\caption{Exhaustive search for the best decision stump. \Algo{BestStump}
  receives a sorted column (feature) $\bs$ of the observation matrix $\bX$. The
  sorting in line~\ref{linStumpSort} can be done once for all features outside
  of the boosting loop. \Algo{BestStump} examines all thresholds $b$ halfway
  between two non-identical coordinates $s_i \not = s_{i+1}$ and returns the
  threshold $b^*$ and vote vector $\bv^*$ that maximizes the edge
  $\gamma(\bv,\varphi_{j,b},\bW)$. \Algo{StumpBase} then sets the coefficient
  $\alpha_j$ according to (\ref{eqnMulticlassAlpha}) and chooses the stump
  across features that minimizes the energy
  (\ref{eqnEnergy}). \label{figStumpBase}} \vspace{2cm}
\end{figure}

\begin{figure}[!ht]
  \centerline{
    \includegraphics[width=0.6\textwidth]{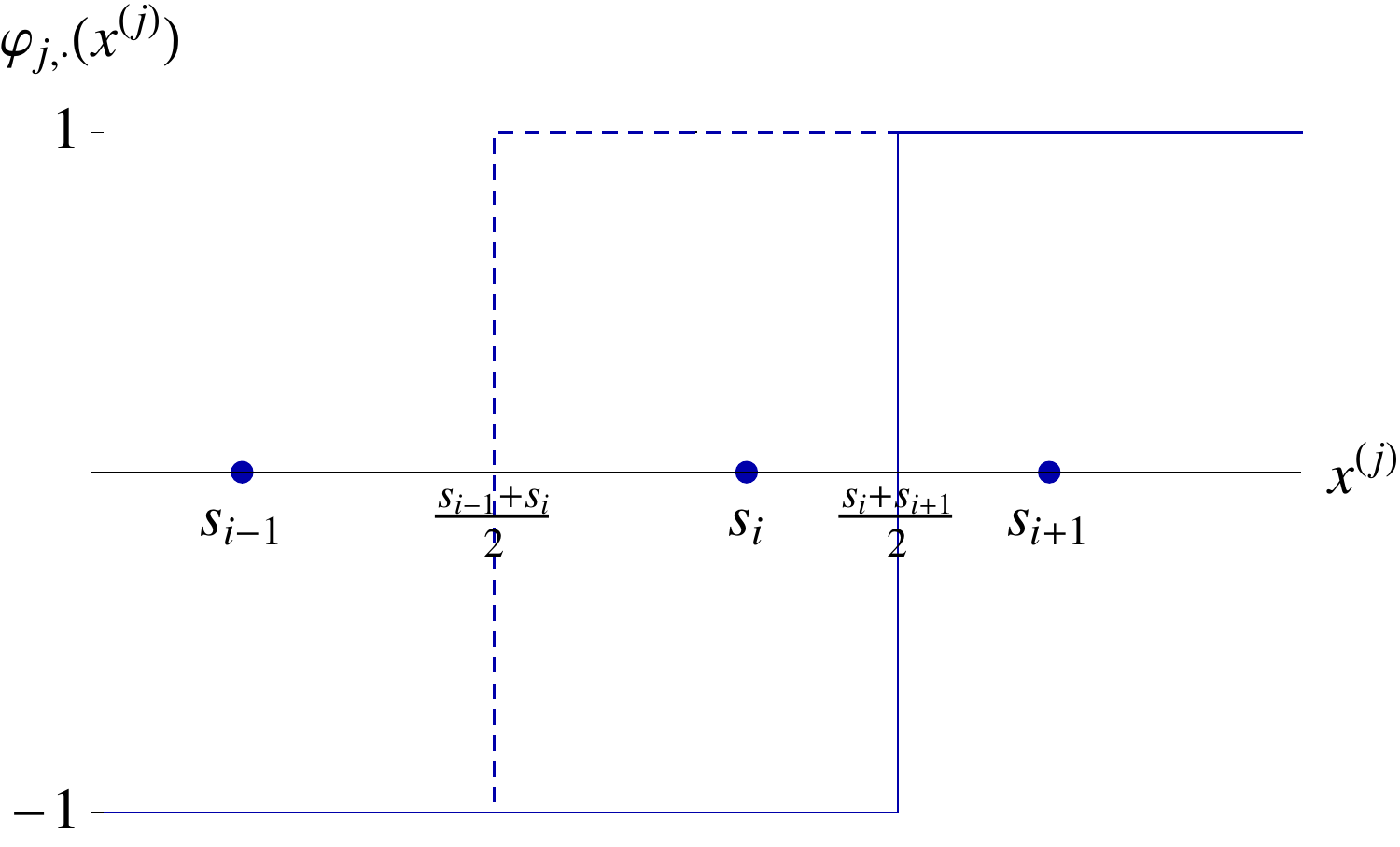}
  }
  \caption{Updating the edge $\gamma_\ell$ in line~\ref{linStumpUpdate} of
  $\Algo{BestStump}$. If $y_{i,\ell}=1$, then $\gamma_\ell$ decreases
  by $2w_{i,\ell}$, and if $y_{i}=-1$, then $\gamma_\ell$ increases
  by $2w_{i,\ell}$.
    \label{figStumpUpdate}}
\end{figure}

\section{Cutting the data set}
\label{secCutDataSetProductBase}

The basic operation when adding a tree node with a scalar binary classifier
(cut) $\varphi$ is to separate the data matrices $\bX$, $\bY$, and $\bW$
according to the sign of the classification $\varphi(\bx_i)$ for all $\bx_i \in
\bX$. Figure~\ref{figCutDataSetProductBase} contains the pseudocode of this
simple operation.

\begin{figure}[!ht]
\centerline{
  \begin{algorithm}{$\Algo{CutDataSet}\big(\bX,\bY,\bW,\varphi(\cdot)\big)$}
    \Aitem $\bX_- \setto \bY_- \setto \bW_- \setto \bX_+ \setto \bY_+ \setto
    \bW_+ \setto ()$ \mt \algoremark{empty vectors}
    \Aitem \For $i  \setto 1$ \To $n$
    \Aitem \mt \If $\bx_i \in \bX$ \Then
    \Aitem \mtt \If $\varphi(\bx_i) = -1$ \Then
    \Aitem \mttt $\bX_- \setto \Algo{Append}(\bX_-,\bx_i)$
    \Aitem \mttt $\bY_- \setto \Algo{Append}(\bY_-,\by_i)$
    \Aitem \mttt $\bW_- \setto \Algo{Append}(\bW_-,\bw_i)$
    \Aitem \mtt \Else
    \Aitem \mttt $\bX_+ \setto \Algo{Append}(\bX_+,\bx_i)$
    \Aitem \mttt $\bY_+ \setto \Algo{Append}(\bY_+,\by_i)$
    \Aitem \mttt $\bW_+ \setto \Algo{Append}(\bW_+,\bw_i)$
    \Aitem \Return $(\bX_-, \bY_-, \bW_-, \bX_+, \bY_+, \bW_+)$
  \end{algorithm}
}
  \caption{The basic operation when adding a tree node is to separate the data
    matrices $\bX$, $\bY$, and $\bW$ according to the sign of classification
    $\varphi(\bx_i)$ for all $\bx_i \in \bX$. \label{figCutDataSetProductBase}}
\end{figure}

\end{document}